\documentclass[conference]{IEEEtran}
\IEEEoverridecommandlockouts
\usepackage{cite}
\usepackage{amsmath,amssymb,amsfonts}
\usepackage{algorithmic}
\usepackage{graphicx}
\usepackage{textcomp}
\usepackage{xcolor}
\usepackage{multirow}

\def\BibTeX{{\rm B\kern-.05em{\sc i\kern-.025em b}\kern-.08em
    T\kern-.1667em\lower.7ex\hbox{E}\kern-.125emX}}
\begin{document}

\title{Squeeze-and-Excitation on Spatial and Temporal Deep Feature Space for Action Recognition\\
}

\author{\IEEEauthorblockN{Gaoyun An, Wen Zhou, Yuxuan Wu, Zhenxing Zheng, Yongwen Liu}
\IEEEauthorblockA{\textit{Institute of Information Science, Beijing Jiaotong University, Beijing 100044, China}\\
\textit{Beijing Key Laboratory of Advanced Information Science and Network Technology, Beijing 100044, China}\\
Email:\{gyan, 16125155, 16120307, zhxzheng, 17120314\}@bjtu.edu.cn}
}

\maketitle

\begin{abstract}
Spatial and temporal features are two key and complementary information for human action recognition. In order to make full use of the intra-frame spatial characteristics and inter-frame temporal relationships, we propose the Squeeze-and-Excitation Long-term Recurrent Convolutional Networks (SE-LRCN) for human action recognition. The Squeeze and Excitation operations are used to implement the feature recalibration. In SE-LRCN, Squeeze-and-Excitation ResNet-34 (SE-ResNet-34) network is adopted to extract spatial features to enhance the dependencies and importance of feature channels of pixel granularity. We also propose the Squeeze-and-Excitation Long Short-Term Memory (SE-LSTM) network to model the temporal relationship, and to enhance the dependencies and importance of feature channels of frame granularity. We evaluate the proposed model on two challenging benchmarks, HMDB51 and UCF101, and the proposed SE-LRCN achieves the competitive results with the state-of-the-art.
\end{abstract}


\section{Introduction}
In recent years, research on human action recognition in images and videos has become a hot topic in the fields of computer vision, pattern recognition, and machine learning \cite{Karpathy2014Large}. In the field of image recognition, features learned by convolutional neural networks (CNN) are often superior to hand-crafted ones \cite{Ng2015Beyond}. Therefore, CNN applies to spatial feature extraction of human action. The temporal relationship between video frames provides additional motion information. Effective usage of the temporal information in the video can better analyze the potential information in the video and improve the recognition rate.

Simonyan et al. \cite{Simonyan2014T} proposed a two-stream CNN network in which RGB and optical flow images were inputted into two separate networks to learn appearance and motion characteristics. The final result was the fusion of the prediction of two streams. However, space and time complexity of the pre-calculation of optical flow is very high. Even when using GPUs, optical flow calculation is the main bottleneck for two-stream CNN network. Ji et al. \cite{Xu20123D} proposed the 3D CNN model, which used a 3D convolution kernel to convolve three consecutive frames and obtained the spatial-temporal characteristics. Donahue et al. \cite{Donahue2016Long} proposed the LRCN model, which is a typical action-aware deep convolutional network composed of AlexNet \cite{Krizhevsky2012ImageNet} and LSTM \cite{Hochreiter1997Long} network. AlexNet extracts the intra-frame spatial features and fed them into an LSTM for modeling temporal relationship to obtains spatial and temporal information.

We propose the Squeeze-and-Excitation Long-term Recurrent Convolutional Networks (SE-LRCN) for human action recognition, which uses the Squeeze-and-Excitation operations to implement the feature recalibration. Therefore, this paper also uses the Squeeze-and-Excitation (SE) module \cite{lei2017sru} to name this model. The framework of the proposed SE-LRCN is shown as in Fig.1. In SE-LRCN, the Squeeze-and-Excitation ResNet-34 (SE-ResNet-34) \cite{lei2017sru} extracts the spatial feature, and we also propose Squeeze-and-Excitation Long Short-Term Memory (SE-LSTM) for modeling the temporal relationship. SE-ResNet-34 and SE-LSTM achieve the attention extraction of pixel and frame granularity respectively. Taking into consideration the dependencies and degrees of importance of the feature channels of the pixel granularity when extracting the spatial features; Considering that the temporal relationship and degree of importance of the frame granularity when to perform temporal modeling. We evaluate the proposed model on two challenging benchmarks, HMDB51 and UCF101, and achieves the competitive results with the state-of-the-art

\begin{figure*}[htbp]
	\centerline{\includegraphics[width=17cm]{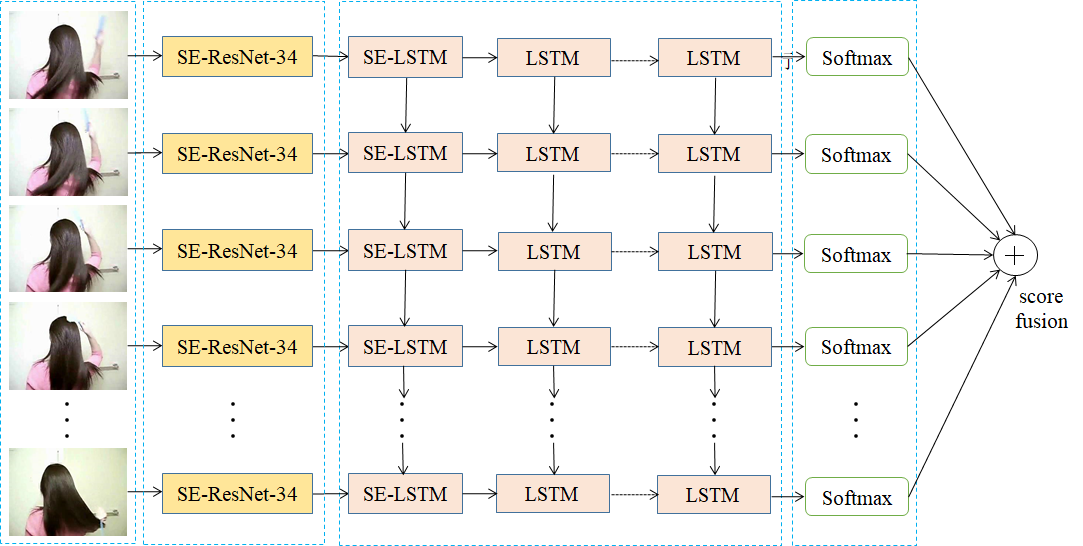}}
	\caption{The framework of SE-LRCN.}
	\label{fig}
\end{figure*}

\section{Related work}
It is very important for improving the accuracy of action recognition to extract spatial and temporal features with high discrimination from human action videos.
\subsection{Hand-crafted features}
Bobick et al. \cite{Bobick2001The} used the moving human body in the video as a spatial-temporal model to present Motion Energy Image (MEI) and Motion History Image (MHI), which could reflect the temporal changes. Wang et al. \cite{Wang2011Action, Wang2014Action} proposed Dense Trajectories (DT) and improved Dense Trajectories (iDT) to extract apparent features and motion information from the detected spatial-temporal salient regions. The commonly used descriptors are pixel intensity, pixel distribution, optical flow characteristics, gradient direction, gradient strength. Laptev et al. \cite{Laptev2008Learning} extracted the Histogram of Oriented Gradient (HOG) and the Histogram of Optical Flow (HOF) features, and combined them into one feature vector HOG/HOF. The HOG/HOF feature describes the appearance and movement information in salient points and regions.
\subsection{Deep-learning features}
Ji et al. \cite{Xu20123D} extended 2DCNN to 3DCNN. Multiple levels of operations such as 3D convolution and pooling are performed on each channel, then the information of each channel is merged to obtain the final feature description of the video, which includes the spatial and temporal characteristics of the video. Karpathy et al. \cite{Karpathy2014Large} proposed the Slow Fusion Model to fuse each frame and use the correlation of neighbor frames in the video to improve the expression ability. Simonyan et al. \cite{Simonyan2014T} proposed Two-Stream ConvNets, which simultaneously constructed both spatial and temporal streams. Spatial stream processes still frame to get spatial and shape information. Time stream processes multiple optical streams stacked in succession to obtain time and motion information. Jeff et al. \cite{Donahue2016Long} proposed the LRCN algorithm, which sent the intra-frame spatial features extracted by AlexNet into LSTM to model temporal relationships. Wang et al. \cite{Wang2016Temporal} combined sparse time sampling strategy and video level supervision to use the entire action video to achieve efficient and effective learning.

\section{Our Approach}
The architecture of SE-LRCN network is shown in Fig.1, which uses the deeper ResNet-34 network as the feature extractor to replace the shallower AlexNet network in LRCN. The proposed SE-LRCN adopts the feature recalibration strategy for intra-frame and inter-frame feature extraction, named as SE-ResNet-34 and SE-LSTM network respectively. Finally, the output value is normalized using the softmax function to get the final prediction. The network contains multiple LSTM layers, and the SE module only applies to the first layer.

\subsection{SE-ResNet-34 feature extraction}
\textbf{Basebone network - ResNet-34}. Some experiments show that deeper features extracted by the network have more semantic information. Therefore, increasing the depth of network is very important to improve performance and enrich layering features. However, the deeper common networks suffer from gradient disappearance and explosion \cite{Glorot2010Understanding}, which can be solved by normalization initialization and intermediate layer standardization. If the layer behind the deep network is the identity mapping, the model will degenerate into a shallow network, which can solve the problem of network degradation in deep networks.

He et al. \cite{He2015Deep} proposed Residual Network (ResNet), which introduced identity mappings on the basis of the original normal deep network, and used the sum of the input and residual block as the final output to solve the degradation problem. In each residual block, shortcut connection between the input and the concatenated output of two or three layers is added. This operation is called identity mappings. The ResNet-34 [8] network consists of thirty-three convolution layers and one fully connected layer. From the bottom layer to the top layer are Conv1, Conv2\_x, Conv3\_x, Conv4\_x, Conv5\_x, and Fc layers, respectively, where Conv1 represents a convolution layer, the convolution kernel size is  , and Conv2\_x, Conv3\_x, Conv4\_x, and Conv5\_x represent residual convolution blocks with a convolution kernel size $3\times3$.

Since the model pre-trained by ImageNet has better image representation capability, this paper uses the ResNet-34 model pre-trained with ImageNet as the backbone network.

\textbf{SE-ResNet-34}. Squeeze-and-Excitation Networks (SENet) \cite{lei2017sru} considered the feature channel relationship. In order to explicitly model the interdependencies among the feature channels and not introduce new spatial dimensions, a "feature recalibration" strategy is used. Specifically, the model automatically acquires the degree of importance of each feature channel through learning methods, and then according to the degree of importance, it enhances the channel features that are useful for the current task and suppresses the useless channel features. It also achieves attention extraction of feature channel granularity.

Here, we perform feature recalibration on the $3^{rd}$ residual block of the Conv\_5x layer of the ResNet-34 network, which is also the last residual block. An SE-ResNet-34 network is constructed for spatial feature extraction, which implements SE operations as follows:

The Squeeze operation is:

\begin{equation}
z_{c} = F_{sq}\left ( u_{c}^{conv\_5x\_3} \right )=\frac{1}{W\times H}\sum_{i=1}^{W}\sum_{j=1}^{H}u_{c}^{conv\_5x\_3}(i,j)
\end{equation}

Where $z_{c}$ is the c-th element of $z\in \boldsymbol{R}^{C \times1}$, C is the number of channels of Conv\_5x layer of the ResNet-34 network, $W\times H$ represents the dimension of each channel. $u_{c}^{conv\_5x\_3}$ is the c-th feature map of the 3rd residual block of Conv\_5x layer. Here, $C=512$ and $W=H=7$.

The Excitation operation is:

\begin{equation}
s = F_{ex}\left ( z,W \right )=\sigma \left(W_{2}\delta \left(W_{1}z \right ) \right )
\end{equation}
Where $\delta$ is a ReLU activation function, $\sigma$ is a Sigmoid function, $W_{1}\in R^{\frac{C}{r}\times C}$ is the fully connected (FC) layer for dimensionality reduction, $W_{2}\in R^{C \times \frac{C}{r}}$is the FC layer for dimensionality increasing, and here $r=16$.

Finally, the new output of Conv\_5x layer is:

\begin{equation}
\widetilde{u}_{c}^{conv\_5x\_3}= u_{c}^{conv\_5x\_2}+s_{c}\cdot u_{c}^{conv\_5x\_3}
\end{equation}

Where $u_{c}^{conv\_5x\_2}$ is the c-th feature map of the $2^{nd}$ residual block of Conv\_5x layer, and the meaning of other symbols is the same with the former ones.

After the global average pooling operation is performed on $\widetilde{u}^{conv\_5x\_3}$, the spatial feature representation $\widetilde{U} \in R^{T \times C\times 1}$ of the video is obtained, where $T$ represents the total number of frames of the video, $C\times 1$ represents the feature vector of the video is globally averaged. Then the three-dimensional feature matrix formed by each video is sent to the SE-LSTM network for temporal relationship modeling.

\subsection{SE-LSTM for temporal relationship modeling}

RNN \cite{Mikolov2011Extensions} is a general term for a series of neural networks that can process time series data. The network has feedback links, which can use past frame information to assist in understanding the current frame. LSTM is a kind of RNN network. For the sake of considering the interdependencies between frames and the importance of different frames, we propose the SE-LSTM network to achieve the attention-grabbing of the frame granularity. The network structure is shown in Fig. 2.

\begin{figure}[htbp]
	\centerline{\includegraphics[width=7cm]{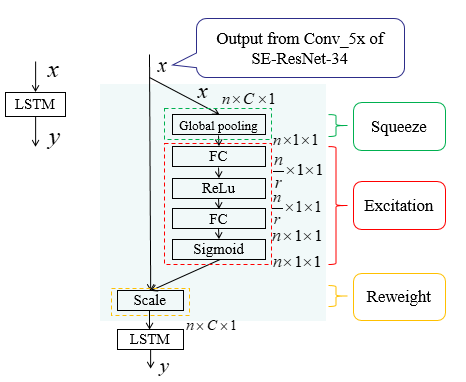}}
	\caption{The structure of the original LSTM module (left) and the SE-LSTM model (right).}
	\label{fig}
\end{figure}

\textbf{SE-LSTM}. In order not to introduce new feature dimensions, we still use the "feature recalibration" strategy to implement the SE-LSTM. The model automatically learns the importance of each frame by learning, and improves the useful features of the current frame and suppresses those of useless frames. The SE-LSTM network implements SE operations as follows:

There are two Squeeze operations to enhance the dependencies and importance of feature channels of frame granularity:
\begin{equation}
z_{t}=F_{sq}\left(\widetilde{U}_{t} \right)=\frac{1}{C}\sum_{c=1}^{C}\widetilde{U}_{t,c} 
\end{equation}
or
\begin{equation}
\widetilde{z}_{c}=F_{sq}\left(\widetilde{U}_{c} \right)=\frac{1}{T}\sum_{t=1}^{T}\widetilde{U}_{t,c} 
\end{equation}

Where $z_{t}$ is the t-th element of $z\in R^{T\times 1}$, $\widetilde{z}_{c}$ is the c-th element of $z\in R^{C\times 1}$, $C$ is the number of channels of Conv\_5x layer of the SE-ResNet-34 network, $T$ represents the total number of frames of the video. In this paper, only formula $(4)$ is used to enhance the dependencies and importance of feature channels of frame granularity. Formula $(5)$ is also suitable to SE-LSTM.

The Excitation and Reweight operations are similar to formula $(2)$ and $(3)$ respectively.

\subsection{Network structure and prediction}
The SE-LRCN network architecture is shown in Fig.1. First, the features of the video are extracted using SE-ResNet-34, and the activation tensor output from the Conv\_5x layer is selected as the feature value. When using a one-layer LSTM network for temporal relationships modeling, SE operation is performed before the LSTM network. When using multi-layer LSTMs, SE operation is only performed on the first layer of LSTMs. The last layer of LSTMs gets the value $\widehat{y}_{t}$, and a softmax layer is adopted to get the prediction result of the t-th frame:
\begin{equation}
P\left(y_{t}=c \right )=softmax\left(\widehat{y}_{t} \right )=\frac{exp\left(\widehat{y}_{t,c} \right )}{\sum _{c'\in C}exp \left(\widehat{y}_{t,c'} \right )}, c\in C
\end{equation}
where, $C$ represents the number of categories.

Finally, late-fusion is used to fuse the predicted values of all frames in the video to obtain the predicted value of the entire video. The method of late-fusion includes the mean late-fusion and the maximum late-fusion. We used the mean late-fusion method.

\section{Experimental evaluation}
\subsection{Dataset}
The HMDB51 dataset \cite{Kuehne2011HMDB} has a total of 6849 samples and is divided into 51 categories. Each category contains at least 101 video samples. The sources of the data set are different, and there are many factors such as video occlusion, camera movement, complex background, and changes in lighting conditions, resulting in low recognition accuracy and challenge. The UCF101 dataset \cite{Soomro2012UCF101} contains 101 types of behavioral categories with a total of 13320 videos (a total of 27 hours). It is one of the data sets with the largest number of action categories and the largest number of samples.

The UCF101 and HMDB51 datasets have three classification criteria for the training set and test set, namely split 1, split 2, and split 3. We used the split 1 of the two datasets as the classification criteria during experiments.

We divide each video into multiple segments, which consists of consecutive 30 frames of video frames. Each segment from the same original video is preceded by 15 frames of coincident frames. Through the coincidence operation, the number of samples of the training set and the test set is increased. If the length of the last segment of the video does not exceed 30 frames, the video is looped for completion. The short side of the original frame is scaled to 256 and horizontally inverted with a 50\% probability. The original or the inverted frame is randomly cropped to size $224\times 224$  which contributes to the purpose of data enhancement. The normalized formula is channel=(channel-mean)/std, and the average mean value of each channel is $(0.485, 0.456, 0.406)$, and std is $(0.229, 0.224, 0.225)$. After normalizing, samples are sent to the network. We use random cropping, and 50\% probability of horizontal inversion on all training set images to achieve data enhancement. The test image uses center cropping and no horizontal inversion.

In this paper, the entire SE-LRCN network is trained using the Adam optimization algorithm \cite{Kingma2014Adam} and cross-entropy loss, The epochs are 16 times, the learning rate is $1e-5$, the learning rate is decreased by 10\% and the batch is set to 28 per iteration. The Dropout value is set to 0.5. ResNet-34 is pre-trained by ImageNet.
\subsection{Algorithm comparison}
\textbf{Improved LRCN algorithm}. Each frame of each video is sent into ResNet-34 to extract spatial features, and the activation tensor output from the Conv\_5x layer is selected as the spatial feature description. A global average pooling operation is performed on the feature description to obtain a new feature vector representation. Then it is sent to LSTM to get the final classification result.

In order to compare the effect of the LSTM layer number and hidden layer neural unit number on the network, we set up experiments with different numbers of hidden units and different LSTM layers. 

The experimental results of the improved LRCN algorithm on the UCF101 and HMDB51 data sets are shown in Table. \ref{p} and Table.  \ref{q}. Comparing the recognition rate results, it can be found that when the number of LSTM layers is fixed, the recognition rate increases as the number of hidden neurons increases. Among them, when the number of LSTM layers is 3 and the number of hidden units is 1024, the highest recognition rate is achieved, which is 80.96\% and 49.41\%, respectively. This experiment verifies the effectiveness of the three-layer LSTM and 1024 hidden units for human behavior recognition research.

\begin{table}[htbp]
	\caption{Comparison of the results of the improved LRCN algorithm with different parameters in the UCF101 dataset}\label{p}
	\begin{center}
		\begin{tabular}{cccc}
			\hline
			setting&2 layers&3 layers&4 layers\\
			\hline
			256 hidden units&78.92&79.90&79.21\\
			512 hidden units&79.56&80.19&79.74\\
			1024 hidden units&80.34&\textbf{80.96}&80.53\\
			\hline
		\end{tabular}
	\end{center}
\end{table}

\begin{table}[htbp]
	\caption{Comparison of the results of the improved LRCN algorithm with different parameters in the HMDB51 dataset}\label{q}
	\begin{center}
		\begin{tabular}{cccc}
			\hline
			setting&2 layers&3 layers&4 layers\\
			\hline
			256 hidden units&47.06&47.71&47.91\\
			512 hidden units&47.84&48.04&48.23\\
			1024 hidden units&48.17&\textbf{49.41}&48.69\\
			\hline
		\end{tabular}
	\end{center}
\end{table}

Comparing the LRCN with the improved LRCN algorithm, we can find that the improved LRCN algorithm in the UCF101 and HMDB 51 datasets is improved by 9.84\% and 8.76\% respectively compared with the LRCN recognition rate, as shown in Table 3. So as the feature extractor, the deeper ResNet-34 network is more effective than the shallower AlexNet network for human action recognition task.

\begin{table}[htbp]
	\caption{Comparison of the recognition accuracy between LRCN and improved LRCN algorithm}
	\begin{center}
		\begin{tabular}{cccc}
			\hline
			model&structure&UCF101&HMDB51\\
			\hline
			LRCN\cite{Donahue2016Long}&AlexNet+LSTM&71.12&--\\
			\textbf{Improved LRCN}&\textbf{ResNet-34+LSTM}&\textbf{80.96}&\textbf{49.41}\\
			\hline
		\end{tabular}
	\end{center}
\end{table}

\textbf{SE-LRCN algorithm}. According to the above experimental results, when the number of LSTM layers is 3 and the number of hidden units is 1024, the highest recognition rate can be obtained. Therefore, in our experiment, we choose 3 layers LSTM with 1024 hidden unit numbers. We conducted four comparative experiments as shown in Table 4.
\begin{table}[htbp]
	\caption{Comparison of the results of the SE-LRCN algorithm with different model}\label{structure}
	\begin{center}
		\begin{tabular}{cccc}
			\hline
			model&structure&UCF101&HMDB51\\
			\hline
			LRCN \cite{Donahue2016Long}&AlexNet+LSTM&71.12&-\\
			\hline
			
			\multirow{3}{*}{\textbf{SE-LRCN}}&SE-ResNet-34+LSTM&81.77&50.71\\
			&ResNet-34+SE-LSTM&81.30&50.19\\
			&\textbf{SE-ResNet-34+SE-LSTM}&\textbf{82.49}&\textbf{50.98\textbf{}}\\
			\hline
		\end{tabular}
	\end{center}
\end{table}

Through comparison experiments, it can be found that when performing SE operations on the ResNet-34 and LSTM models respectively, the experiments on both data sets are improved, and the highest recognition rate of the SE-ResNet-34+SE-LSTM model is 82.49 \% and 50.98\%, 1.53\% and 1.57\% higher than the ResNet-34+LSTM model respectively, 11.37\% and 10.33\% higher than the AlexNet+LSTM model respectively. This result shows that the SE-ResNet-34+SE-LSTM model could effectively achieve the attention extraction of pixel and frame granularity respectively.

%\begin{figure}[htbp]
%	\centerline{\includegraphics{UCF_barnew.jpg}}
%	\caption{Confusion matrix of SE-LRCN model on the UCF 101databset.}
%	\label{fig}
%\end{figure}

For the SE-LRCN (SE-ResNet-34+SE-LSTM) model,  the recognition rate results for each class of HMDB51 datasets are shown in Fig. 3. In the figure, the horizontal axis represents 51 categories, and the vertical axis represents the recognition rate result corresponding to each category. It can be found that the glof, pullup, ride\_bike and other categories have a better recognition rate, but the recognition rate of the jump, pickup, swing\_baseball, sword\_exercise, hit and other types is poor.

\begin{figure*}[htbp]
	\centerline{\includegraphics[width=15cm]{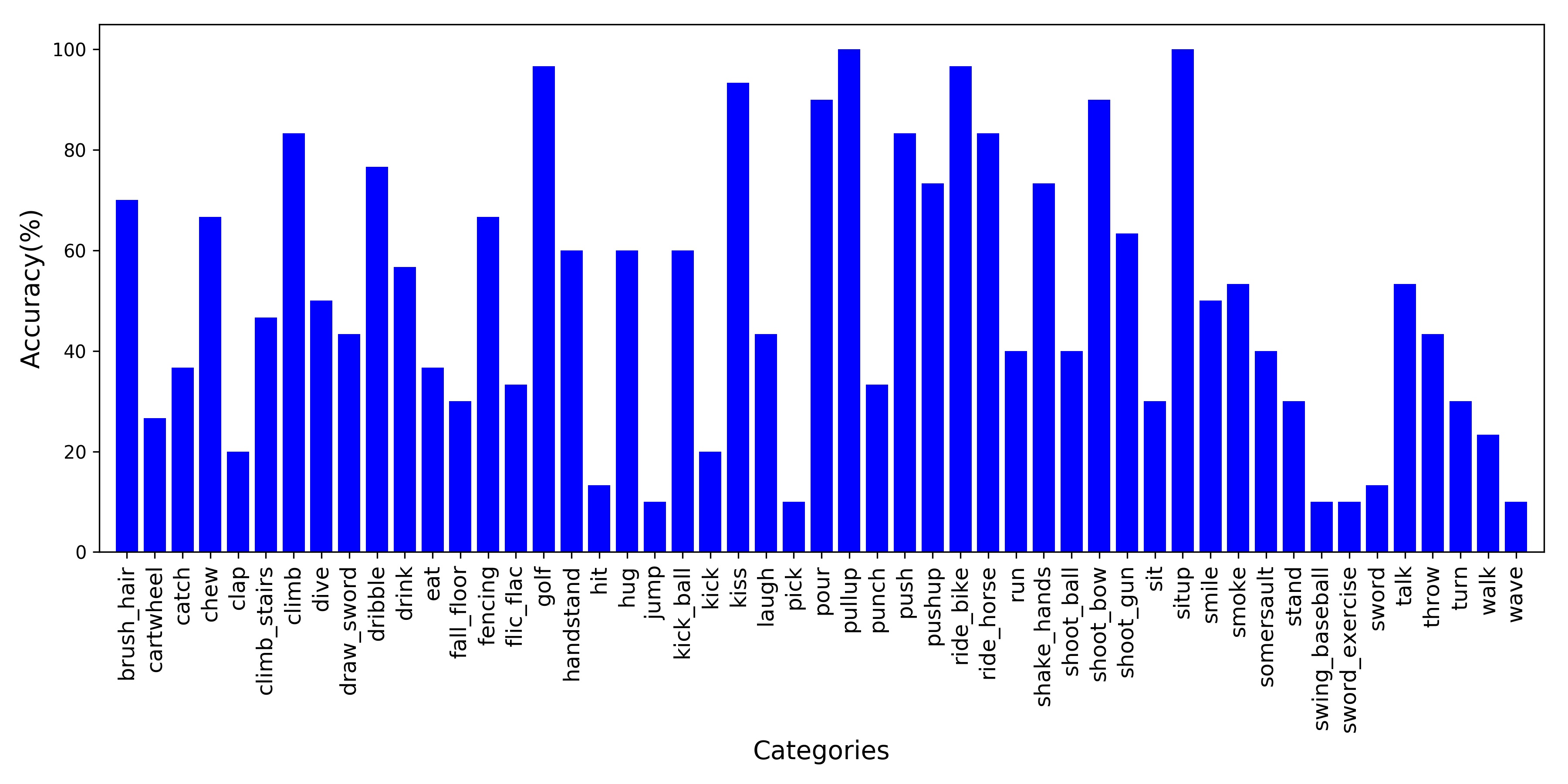}}
	\caption{Per-class recognition rate results of SE-LRCN model on the HMDB 51 dataset.}
	\label{fig}
\end{figure*}

%\begin{figure}[htbp]
%	\centerline{\includegraphics{HMDB_confusion.jpg}}
%	\caption{Confusion matrix of SE-LRCN model on the UCF 101databset.}
%	\label{fig}
%\end{figure}

%\begin{figure}[htbp]
%	\centerline{\includegraphics{HMDB_barnew.jpg}}
%	\caption{Confusion matrix of SE-LRCN model on the UCF 101databset.}
%	\label{fig}
%\end{figure}

\textbf{State-of-the-art analysis}. In this section, we compare the proposed SE-LRCN with human action recognition algorithms in other literature on the UCF101 and the HMDB51 dataset, as shown in Table 5.

\begin{table}[htbp]
	\caption{Comparisons with State-of-the-art}
	\begin{center}
		\begin{tabular}{ccc}
			\hline
			model&UCF101&HMDB51\\
			\hline
			iDT+HD\cite{Wang2014Action}&84.7&57.2\\
			Hybrid representation\cite{Peng2016Bag}&\textbf{87.9}&\textbf{61.1}\\
			\hline
			Slow Fusion Network \cite{Karpathy2014Large}&65.4&--\\
			Spatial ConvNet \cite{Simonyan2014T}&73.0&40.5\\
			Soft Attention \cite{Sharma2015Action}&--&41.3\\
			C3D Model\cite{D2015Learning}&82.3&--\\
			LRCN\cite{Donahue2016Long}&71.12&--\\
			\textbf{SE-LRCN}&\textbf{82.49}&\textbf{50.98}\\
			\hline
			Temporal ConvNet\cite{Simonyan2014T}&83.7&--\\
			Two-Stream\cite{Simonyan2014T}&88.0&59.4\\
			TDD+iDT\cite{wang2015action}&91.5&65.9\\
			TSN\cite{Mikolov2011Extensions}&\textbf{94.2}&\textbf{69.4}\\
			\hline
		\end{tabular}
	\end{center}
\end{table}

These algorithms are divided into three parts for comparison. The first part is algorithms based on hand-crafted features. The second part is algorithms based on deep learning using only RGB data as the model input. The third part is algorithms based on deep learning using both RGB and optical flow data as input.

Comparing within the second part, we can find that SE-LRCN achieved a better recognition rate. This shows the benefit and effectiveness of Squeeze-and-Excitation both on spatial and temporal deep feature space. The recognition rate of SE-LRCN is obviously better than that of Spatial ConvNet\cite{Simonyan2014T}, it is because Spatial ConvNet only learns the feature representation in spatial space. However, SE-LRCN extracts not only spatial information but also temporal information. It shows that extra temporal motion information is beneficial to action recognition. The 3D convolutional network \cite{wang2015action} extracts spatial-temporal features from multiple adjacent frames, the recognition rate of which is slightly lower than that of SE-LRCN.

According to the first and second parts, the recognition rate of the deep learning model using only RGB images as input is lower than that of hand-crafted features. The reason for this is that the training data of the model is limited, so the deep learning model could not learn efficiently.

Comparing the second part with the third one, we found that the recognition rate of SE-LRCN is lower than those of algorithms in the third part. By analyzing algorithms in the third part, it can be found that when the optical flow and RGB information are fused, the recognition rate can be improved by a large margin. SE-LRCN  uses RGB images as input, and it could also take optical flow as input. However, space and time complexity of the pre-calculation of optical flow is very high. In addition, optical flow is very sensitive to light, which will affect the recognition effect. Therefore, it is more difficult to apply optical flow in most real applications.

\section{Conclusion}
In order to enhance the dependencies and importance of the feature channels of frame granularity, the SE-LSTM network is proposed. Based on feature recalibration strategy for intra-frame and inter-frame feature extraction, the SE-LRCN model is proposed for human action recognition. The proposed SE-LRCN could model the spatial and temporal relationships while taking into account the importance of both the pixel feature channel and video frames. So the intra-frame and inter-dependency relationships of video behaviors are discovered by constructing attentional mechanisms on the granularity of pixels and frames. We evaluate the proposed model on two challenging benchmarks, HMDB51 and UCF101, and achieves the competitive results with the state-of-the-art.

\section*{Acknowledgment}

This work was supported partly by the fundamental research funds for the central universities (K17JB00390), the National Natural Science Foundation of China (61772067, 61472030).

\end{document}